\documentclass[a4paper]{article}
\usepackage[dvipsnames]{xcolor}

\newcommand{\remove}[1]{}
\newcommand{\embosi}{Mboshi}
\newcommand{\ligaikuma}{Lig\_Aikuma}

\usepackage{INTERSPEECH2018}
\usepackage[OT1]{fontenc}
\usepackage[utf8]{inputenc}
\usepackage{hyperref}

\title{Unsupervised Word Segmentation from Speech with Attention}
\name{Pierre Godard$^{\ast}${$^{1}$}, Marcely Zanon Boito$^{\star}${$^{1}$}, Lucas Ondel$^{\diamond}$, Alexandre Berard$^{\star}{^{\ddagger}}$, \linebreak
François Yvon$^{\ast}$, Aline Villavicencio$^{\dagger}$, Laurent Besacier$^{\star}$}

\address{
          $^{\ast}$LIMSI, CNRS, Université Paris-Saclay, Orsay, France \linebreak
          $^{\star}$LIG, UGA, G-INP, CNRS, INRIA, Grenoble, France\\
          $^{\diamond}$BUT, Brno, Czech Republic \linebreak
          $^{\dagger}$CSEE, University of Essex, UK \linebreak
          $^{\ddagger}$CRISTAL, Université de Lille, France \linebreak
          }

\email{contact: pierre.godard@limsi.fr, marcely.zanon-boito@univ-grenoble-alpes.fr \\
(1) Both first authors have contributed equally to this paper\\
}

\begin{document}

\maketitle
\begin{abstract}
We present a first attempt to perform attentional word segmentation directly from the speech signal, with the final goal to automatically identify lexical units in a low-resource, unwritten language (UL).
Our methodology assumes a pairing between recordings in the UL with translations in a well-resourced language. It 
uses Acoustic Unit Discovery (AUD) 
to convert speech 
into a sequence of pseudo-phones that is 
segmented using neural soft-alignments produced by a neural machine translation model. 
Evaluation uses an actual Bantu UL, \embosi; 
comparisons to monolingual and bilingual baselines 
illustrate the potential of attentional word segmentation for language documentation.
\end{abstract}


\section{Introduction}

%
%

Speech technology often relies on minimal linguistic expertise and textual information to build acoustic and language models. 
However, for many languages of the world,  text transcripts are limited or nonexistent; therefore, recent 
efforts have been devoted to 
\textit{Zero Resource Settings} \cite{glass2012,jansen2013JHU,zrc2017} 
where the aim is to build speech systems without textual or linguistic resources for e.g.: 
(1) 
unwritten languages 
\cite{besacier_2006, duong2016attentional}; (2) 
models that mimic child  language development \cite{dupoux2016}; (3) documentation of endangered languages by analyzing speech recordings using automatically discovered linguistic units (phones, morphs, words, etc) \cite{adda2016breaking}.


This paper focuses on unsupervised word segmentation 
from speech: the system must output time-stamps delimiting stretches of speech, associated with class labels, corresponding to real words in the language. This task is already considered in the Zero Resource Speech Challenge\footnote{http://zerospeech.com/2017} in a fully unsupervised setting: systems must learn to segment from a collection of raw speech signals only. We investigate here a slightly more favorable case where speech utterances are multilingually grounded: 
they are aligned, at the sentence level, to a written translation in another language
. Such a condition is realistic in language documentation, where it is common to collect speech in the language of interest and have it translated or glossed in another language \cite{blachon2016parallel}. 
 In this context, we want to examine whether
we can take advantage of the weak form of supervision available in these translations to help word segmentation from speech.  Our hypothesis is that textual translations should help in segmenting speech into words in the unwritten language, even in the absence of (manually obtained) phonetic labels.
As a first contribution in this direction, we recently proposed to leverage attentional encoder-decoder approaches for unsupervised word segmentation \cite{boito2017asru}. However, this was done from an unsegmented sequence of (manually obtained) true phone symbols, not from speech. It was shown that the approach proposed can compete with a Bayesian Non Parametric (BNP) baseline \cite{Goldwater09bayesian}  on a small corpus in {\embosi} language (5k sentences only). 

In this paper, our \textbf{contribution} is to develop an attentional encoder-decoder word segmentation from speech ($\S$\ref{sec:approach}) that operates in two steps:
(a) automatic Acoustic Unit Discovery (AUD), based on Bayesian models, to generate time-marked pseudo-phone symbols from the speech; (b) encoder-decoder word segmentation using these pseudo-phones.\footnote{End-to-end speech processing can be performed with an encoder-decoder architecture for speech translation  (e.g.\ \cite{berard2016listen,weiss2017sequence});
early attempts to train end-to-end from speech to translated text, in our language documentation scenario, were not viable due to limited data.}
Experiments with AUD outputs of increasing complexity (see $\S$\ref{sec:AUD}) are presented for word boundary detection using the {\embosi} corpus recently made available \cite{godard2018low} ($\S$\ref{sec:experiments}). 
Our best pipeline from speech has a word boundary F-measure of 50.0\% while segmenting from true phone symbols leads to 61.0\%.

\section{Attentional Encoder-Decoder Approach for Word Discovery}
\label{sec:approach}

For word segmentation, given a parallel corpus pairing sequences of pseudo-phone units in the unwritten language (UL) with sequences of words in the well-resourced language (WRL), we  compute attention matrices as the result of training a standard Neural Machine Translation (NMT) system translating from the WRL into the UL. Then, these soft-alignment 
matrices are  post-processed to derive word boundaries.
 

\subsection{Neural Architecture}\label{subsec:model}

The NMT architecture, equations \eqref{eq:nmt-1}-\eqref{eq:nmt-4},
are inspired by \cite{bahdanau2014neural}. A bidirectional encoder reads the input sequence $x_1,...,x_A$ and produces a sequence of
encoder states $\mathbf{h} = h_1,...,h_A  \in \mathbb{R}^{2\times n}$, where $n$ is the chosen encoder cell size. At each time step $t$,  the decoder uses its current state $s_{t-1}$ and an attention mechanism 
to compute a probability distribution $y_t$ over a target vocabulary of size $|V|$. It then generates the symbol $z_t$ having the highest probability, stopping upon generating a special end-of-sentence token.
The decoder updates its internal representation $s_t$, using the ground-truth symbol $w_t$, instead of the generated symbol $z_t$, since in our alignment setting the reference translations are always available, even at test time. 
Our system is described by the following equations:
\begin{align}
    c_t={\rm attn}(\mathbf{h}, s_{t-1}) \label{eq:nmt-1} \\
    y_t=\operatorname{output}(s_{t-1} \oplus E(w_{t-1}) \oplus c_t) \label{eq:nmt-2}\\
    z_t=\arg \max{y_t} \label{eq:nmt-3}\\
    s_t=\operatorname{LSTM}(s_{t-1}, E(w_t) \oplus c_t),\label{eq:nmt-4}
\end{align}
	
\noindent where $\oplus$ is the concatenation operator. $s_0$ is initialized with the last state of the encoder (after a non-linear transformation), $z_0 = \texttt{<BOS>}$ (special token), and $E\in\mathbb{R}^{|V|\times n}$ is the target embedding matrix.
The $\operatorname{output}$ function uses a maxout layer, followed by a linear projection to $\mathbb{R}^{|V|}$, as in \cite{bahdanau2014neural}.

The attention mechanism is defined as:
\begin{align}
    \label{eq:global}
    e_{t,i} = v^T\tanh{(W_1 h_i + W_2s_{t-1} + b_2)} \\
    \alpha_{t,i} = \operatorname{softmax}(e_{t,i}) \label{eq:attn} \\
	c_t = {\rm attn}(\mathbf{h}, s_{t-1}) = \sum_{i=1}^{A}\alpha_{t,i} h_i
\end{align}
\noindent where $v$, $W_1$, $W_2$, and $b_2$ are learned jointly with the other model parameters.
At each time step ($t$) a score $e_{t,i}$ is computed for each encoder state $h_i$, using the current decoder state $s_{t-1}$. These scores are then normalized 
using a $\operatorname{softmax}$ function, thus giving a probability distribution over the input sequence $\sum_{i=1}^{A} \alpha_{t,i} = 1$ and $\forall t,i, 0 \leq \alpha_{t,i} \leq{} 1$.
The context vector $c_t$ used by the decoder is a weighted sum of the encoder states. This can be understood as a summary of the useful information in the input sequence for the generation of the next output symbol $z_t$. Likewise, 
the weights $\alpha_{t,i}$ can be viewed as defining a soft-alignment between the input $x_i$ and output $z_t$.

\subsection{Word Segmentations from Attention}\label{subsec:approach}

The main aspects of our approach are detailed below.
 
\textbf{Reverse Architecture:} in NMT, the soft alignments probabilities are normalized for each \emph{target symbol} $t$ (i.e.\ $\forall t, \sum_i \alpha_{i,t} = 1$, with $i$ indexing the source symbols). However, there is no similar constraint for the source symbols, as discussed by~\cite{duong2016attentional}. Rather than enforcing additional constraints on the alignments, as in the latter reference, we propose to reverse the architecture and to translate from WRL words into UL symbols, following 
\cite{boito2017asru}.
This ``reverse" architecture notably prevents the attention model from ignoring some UL symbols. As experiments with actual phone sequences have shown that the best results were obtained with this WRL-to-UL translation \cite{boito2017asru}, we will use this reverse architecture throughout.

\textbf{Alignment Smoothing:} to deal with the length discrepancy between UL (pseudo-phones) and WRL (words), 
we implemented the alignment smoothing procedure proposed by \cite{duong2016attentional}. It consists of 
first adding temperature to the \textit{softmax} function (we use T=10 for all experiments) used by the attention mechanism; and then post-processing the resulting soft-alignment probability matrices, averaging each score with the scores of the two neighboring words.
Even if boosting many-to-one alignments should not hold in the case of the reverse architecture, we keep it for our experiments given the gains reported by \cite{boito2017asru}, even in the reverse case.

\textbf{Hard Segmentation Generation:} once the soft-alignment matrices $\mathbf{\alpha}$ are obtained for all utterances in the corpus, a word segmentation is 
inferred as follows. We first transform soft-alignments into hard-alignments by aligning each UL symbol $w_t$ with the word $x_i$ such that: $i=\arg\max_{i'} \alpha_{t,i'}.$ 
The source sequence is then segmented according to these hard-alignments: if two consecutive symbols are aligned with the same WRL word, they are considered to belong to the same UL word.

\section{Acoustic Unit Discovery (AUD)}
\label{sec:AUD}

Our AUD systems are based on the  
Bayesian non-parametric Hidden Markov Model (HMM) of \cite{ondel2016variational}.
This model is topologically equivalent to a phone-loop where each 
acoustic unit is represented by a left-to-right HMM. To cope with the unknown number of units 
needed to properly describe the speech, the model assumes a potentially infinite number of symbols. However, the prior over the weight of the acoustic units (a Dirichlet Process 
\cite{TehJor2010}) will act as a sparsity regularizer, leading to a model which explains the data with a relatively small unit set. 

%
%
We implemented two variants of this original model. The first one, referred to as \textit{HMM}, 
approximates the Dirichlet Process prior by a simpler symmetric Dirichlet 
prior, as proposed by \cite{Kurihara2007}. 
This approximation, while retaining the sparsity constraint, avoids the complication of dealing with the variational
treatment of the stick breaking process in Bayesian non-parametric models. The second variant, denoted Structured Variational AutoEncoder (\textit{SVAE}) AUD, is based on the work of \cite{Johnson2016} and embeds the HMM model 
into a Variational Auto-Encoder (VAE) \cite{KingmaW2013} where the posterior distribution of the HMM and the VAE parameters 
are trained jointly using Stochastic Variational Bayes \cite{Hoffman2013, Johnson2016}. To initialize the 
model, the prior distribution over the HMM parameters (mixture weights, means and covariance matrices) was set to the 
posterior distribution of the phone-loop trained in a supervised fashion (Baum-Welch training) on the TIMIT data set. This 
procedure can be seen as a cross-lingual knowledge transfer as the AUD training on the UL language is essentially adapting 
the English phone set distribution to the {\embosi} corpus. Finally, both models were trained using two features
sets: the well-known \textit{MFCC} + $\Delta$ + $\Delta \Delta$ features and the Multilingual BottleNeck (\textit{MBN}) features 
\cite{Grezl2014}. Note that the MBN features were not trained on any {\embosi} data, and only use languages as listed in \cite{Grezl2014}).

\begin{figure*}[ht]
      \centering
      \small
      \begin{tabular}{lcl} \midrule
      Mboshi & & \texttt{wáá ngá iwé léekundá ngá sá oyoá lendúma saa m ótéma} \\
      French & & \texttt{si je meurs  enterrez-moi dans la forêt oyoa avec une guitare sur la poitrine} \\
     \midrule
      \end{tabular}
      \caption{A tokenized and lowercased sentence pair example in our Mboshi-French corpus.}
      \label{fig:example}
\end{figure*}

\section{Word Segmentation Experiments}
\label{sec:experiments}
\subsection{Corpus, Baselines and Metric}
\label{subsec:database}
We used the Mboshi5k corpus \cite{godard2018low} 
in all our experiments.\footnote{The dataset is  documented in \cite{godard2018low} and available at \url{https://github.com/besacier/mboshi-french-parallel-corpus}} 
{\embosi} (Bantu C25) is a typical Bantu language 
spoken in Congo-Brazzaville.
It is one of the languages documented by the BULB (Breaking the Unwritten Language Barrier) project~\cite{adda2016breaking}. 
This speech dataset was collected following a real language documentation scenario, using  \ligaikuma,\footnote{\url{http://lig-aikuma.imag.fr}} a mobile app specifically dedicated to fieldwork language documentation, which works both on Android powered smartphones and tablets \cite{blachon2016parallel}.
The corpus is multilingual (5,130 {\embosi} speech utterances aligned to French text) and contains  linguists' transcriptions in {\embosi}
in the form of a non-standard graphemic form close to the language phonology. Correct word segmentation of the {\embosi} transcripts was also provided by the linguists and a forced-alignment between speech and  transcripts was computed to obtain time-stamps-delimited word tokens for evaluation. The corpus is split in two parts (\textit{train} and \textit{dev}) for which we give basic statistics in Table~\ref{table:basic-stats}. We also include an example of a sentence pair from our corpus in Figure~\ref{fig:example}.

\begin{table}
  \centering
  \small
  \begin{tabular}{lcrrr}  
  \toprule
  language  & split  & \#sent & \#tokens & \#types \\ \midrule
  Mboshi    & train & 4,616 & 27,563    & 6,196 \\
            & dev   & 514   & 2,993     & 1,146 \\ \midrule
  French    & train & 4,616 & 38,843    & 4,927 \\
            & dev   & 514   & 4,283     & 1,175 \\
  \bottomrule
  \end{tabular}
  \caption{Corpus statistics for the \embosi{} corpus}
  \label{table:basic-stats}
\end{table}

 Our neural (\textit{attentional}) word segmentation is compared with two baselines: a naive bilingual baseline (\textit{proportional}) that segments the source according to the target as if the alignment matrix between symbols (AUD symbols in {\embosi} and graphemes in French) was diagonal;\footnote{Blank spaces on the French side are then used to segment the {\embosi} input.} a monolingual baseline \cite{Goldwater09bayesian} which implements a Bayesian non-parametric approach, where (pseudo)-words are generated by a bigram model over a non-finite inventory, through the use of a Dirichlet process (refered to as \textit{dpseg}). 
We evaluate with the \textit{Boundary} metric from the \textit{Zero Resource Challenge 2017} \cite{ludusan2014,zrc2017}. It measures the quality of a word segmentation and the discovered boundaries with respect to a gold segmentation (P, R and F-score are computed).

\begin{table*}[ht]
\centering
\begin{tabular}{llcccccccccccc}
\toprule
\multicolumn{1}{c}{\textbf{\begin{tabular}[c]{@{}c@{}}AUD\\ feat.\end{tabular}}} & \multicolumn{1}{c}{\textbf{\begin{tabular}[c]{@{}c@{}}AUD\\ model\end{tabular}}} & \multicolumn{3}{c}{\textbf{\begin{tabular}[c]{@{}c@{}}dpseg\\ (monoling.)\end{tabular}}} & \multicolumn{3}{c}{\textbf{\begin{tabular}[c]{@{}c@{}}proportional\\ baseline (biling.)\end{tabular}}} & \multicolumn{3}{c}{\textbf{\begin{tabular}[c]{@{}c@{}}attentional\\ (biling.)*\end{tabular}}} & \multicolumn{3}{c}{\textbf{\begin{tabular}[c]{@{}c@{}}att. average\\ (biling.)+\end{tabular}}} \\
\cline{3-14} 
& \multicolumn{1}{c}{} & \textbf{P} & \textbf{R} & \textbf{F} & \textbf{P} & \textbf{R} & \textbf{F} & \textbf{P} & \textbf{R} & \textbf{F} & \textbf{P} & \textbf{R} & \textbf{F} \\ \hline
\textbf{MFCC} & \textbf{HMM} & 27.9 & 80.2 & 41.3 & 42.6 & 49.9 & 46.0 & 51.6 & 44.9 & 48.0 & 55.5 & 43.7 & 48.9\\ \hline
\textbf{MFCC} & \textbf{SVAE} & 29.8 & 69.1 & 41.7 & 42.2 & 51.9 & 46.6 & 52.7 & 45.0 & 48.5 & 55.7 & 44.1 & 49.2\\ \hline
\textbf{MBN} & \textbf{HMM} & 27.8 & 72.6 & 40.2 & 42.5 & 48.1 & 45.2 & 50.8 & 44.5 & 47.4 & 54.1 & 42.9 & 47.8 \\ \hline
\textbf{MBN} & \textbf{SVAE} & 30.0 & 72.9 & 42.5 & 42.5 & 51.6 & 46.6 & 57.2 & 43.0 & 49.1 & 60.6 & 42.5 & 50.0 \\ \hline\hline
\multicolumn{2}{c}{\textbf{true phones}} & 53.8 & 83.5 & 65.4 & 44.5 & 62.6 & 52.0 & 60.5 & 59.9 & 60.3 & 62.8 & 59.3 & 61.0 \\ \bottomrule

\end{tabular}
\caption{Precision, Recall and F-measure on word boundaries over the Mboshi5k corpus, using different AUD to extract pseudo-phones from speech. True phones toplines are also provided. *averaged scores over 5 different runs; +averaged 5 attention matrices}

\label{tab:word-boundary}
\end{table*}

\subsection{Details of the NMT system}

We use the LIG-CRIStAL NMT system.\footnote{See \url{https://github.com/eske/seq2seq}.} Our models are trained using the Adam algorithm, with a learning rate of $0.001$ and a batch size ($N$) of $32$. We minimize the cross-entropy loss between the output probability distribution $p_t=softmax(y_t)$ and a reference translation $w_t$.
Our models use global attention and bidirectional layer in the encoder; encoder and decoder have 1 layer each, with a cell size of~64. Dropout is applied with a rate equal to~0.5. 
For NMT training, we split the 5,130 sentences into training and development, with about 10\% of the corpus for the latter. However, the soft-alignment matrices are obtained from both train and dev sets after forced-decoding and segmentation is evaluated on all 5,130 utterances.


\subsection{Results}


Unsupervised word segmentation results obtained from speech with different AUD configurations
as well as from true phones (upper-bound performance corresponding to a \textit{topline}) are reported in Table~\ref{tab:word-boundary}, using the \textit{Boundary} metric. 
We trained 5~different NMT models changing the train/dev split\footnote{The difference between best and worst configurations varied from 0.5\% to 1.3\% for AUD, and 1.6\% for true phones.} and either (i) averaging the scores over the 5 runs (columns \textit{att. (biling.)} in Table~\ref{tab:word-boundary}) or (ii) averaging the obtained soft-alignment matrices (columns \textit{att. average} in Table~\ref{tab:word-boundary}). The latter slightly boosts boundary detection performance. 
For all AUD configurations, our method outperforms two baselines (\textit{dpseg} and \textit{proportional}), as well as a pure speech-based baseline using segmental DTW~\cite{aren}, which only achieves a F-score of 19.3 on our data. 
While competitive with true phones, 
the results of the monolingual method (\textit{dpseg}) are heavily degraded on discovered (noisier) units, as also reported by \cite{jansen2013JHU}.
Conversely, our method is much more robust 
to noise and seems better suited for real-world scenarios.
While straightforward, the bilingual baseline (\textit{proportional}) is rather strong compared to its monolingual counterpart (\textit{dpseg}). This suggests that multilingual grounding provides a useful signal for word segmentation from speech. 


Regarding AUD specifically, we observe that the best F-score for word boundary detection was obtained with MBN features and the SVAE model.
The results of our \textit{attentional} segmentation are the best results reported so far on this corpus. 
This confirms that we can effectively take advantage of the weak supervision available in the translations in order to help word segmentation from speech.
 
\subsection{Discussion}

The NMT system requires a sequence of unsegmented symbols (the phones) and their aligned sentence translations in order to provide segmentation. Therefore, the AUD method chosen to encode the speech input has an impact on the quality of the final segmentation. 
Our best word segmentation results (see Table~\ref{tab:word-boundary}) are obtained using the SVAE model (this holds for both Bayesian and neural segmentation approaches). One natural explanation would be to posit that phone boundaries (and consequently word boundaries) are more accurately detected by the SVAE model than the HMM model. \cite{ondel2017bayesian} show that this is true in terms of precision for phone boundaries, and in term of normalized mutual information, but that the recall on these boundaries is lower than its HMM counterpart. This indicates that the SVAE model extract more consistent pseudo-phone units, although it misses some boundaries, than the HMM model, and we confirm here the result of \cite{ondel2017bayesian} showing that this is beneficial for the word segmentation task.

Another additional explanation might be that shorter sequences of symbols are easier to segment. For instance, even if the attention helps the system to better deal with long sentences, it is still prone to performance degradation when faced with very long  sequences of symbols~\cite{bahdanau2014neural}.
Table~\ref{table:pseudo} (left side) shows how the different AUD approaches are encoding the UL sentences.
We observe that the HMM model uses more symbols to represent an utterance, while the SVAE model offers a more concise representation.


\begin{table}
\centering
\begin{tabular}{lccc|ccc}
\toprule
\multicolumn{1}{c}{\textbf{}} & \multicolumn{3}{c|}{\textbf{\begin{tabular}[c]{@{}c@{}}Phones per \\ Sentence\end{tabular}}} & \multicolumn{3}{c}{\textbf{\begin{tabular}[c]{@{}c@{}}Tokens per \\ Sentence\end{tabular}}} \\ \midrule
\textbf{} & avg & max & min & avg & max & min \\ \midrule
\textbf{true phones} & 21.8 & 60 & 4 & 6.0 & 21 & 1 \\ \midrule
\textbf{MFCC HMM} & 37.0 & 95 & 11 & 3.6 & 22 & 1 \\ 
\textbf{MFCC SVAE} & 26.3 & 73 & 7 & 7.6 & 26 & 1 \\
\textbf{MBN HMM} & 32.1 & 93 & 12 & 5.0 & 14 & 1 \\ 
\textbf{MBN SVAE} & 23.4 & 71 & 7 & 5.4 & 21 & 1 \\ \bottomrule 
\end{tabular}
\caption{AUD methods differ in their ability to encode speech utterances (left side); which impacts the final segmentation of the attentional model (right side).}
\label{table:pseudo}
\end{table}

Table~\ref{table:pseudo} also reports information regarding the generated segmentation using a single attention model (right side), showing that our best model (MBN SVAE) results in segmentations that relate closely to the \textit{topline} in terms of number of tokens per sentence.
This best model also achieved a vocabulary size close to the \textit{topline}, of 14,837 types compared to 13,878. This is another clue as to why the MBN SVAE is performing best on our task.


Analyzing the averaged attention model's results in Table~\ref{tab:word-boundary}, we can see an  increase in performance 
of about 0.8\% in all cases.
This improvement also holds for tokens and types scores (not reported here). However, while the \textit{topline} achieves 34.3\% of vocabulary (types) retrieval, our best AUD setup achieves 13.5\% only. 
This illustrates the difficulty of the word discovery task -- a task already challenging with true phones -- in noisy setups.
The large difference between true phones and pseudo-phones for type's retrieval could be explained by the fact that a single change in the pseudo-phone sequence representing two speech segments of a same word will have the consequence to split the word cluster in two parts (in two types). A deeper analysis of the word clusters obtained is probably necessary to better understand how AUD from speech affects the word discovery task, and to come up with 
ways to better cluster speech segments in relevant types.

\begin{figure}
\centering
\includegraphics[scale=0.2]{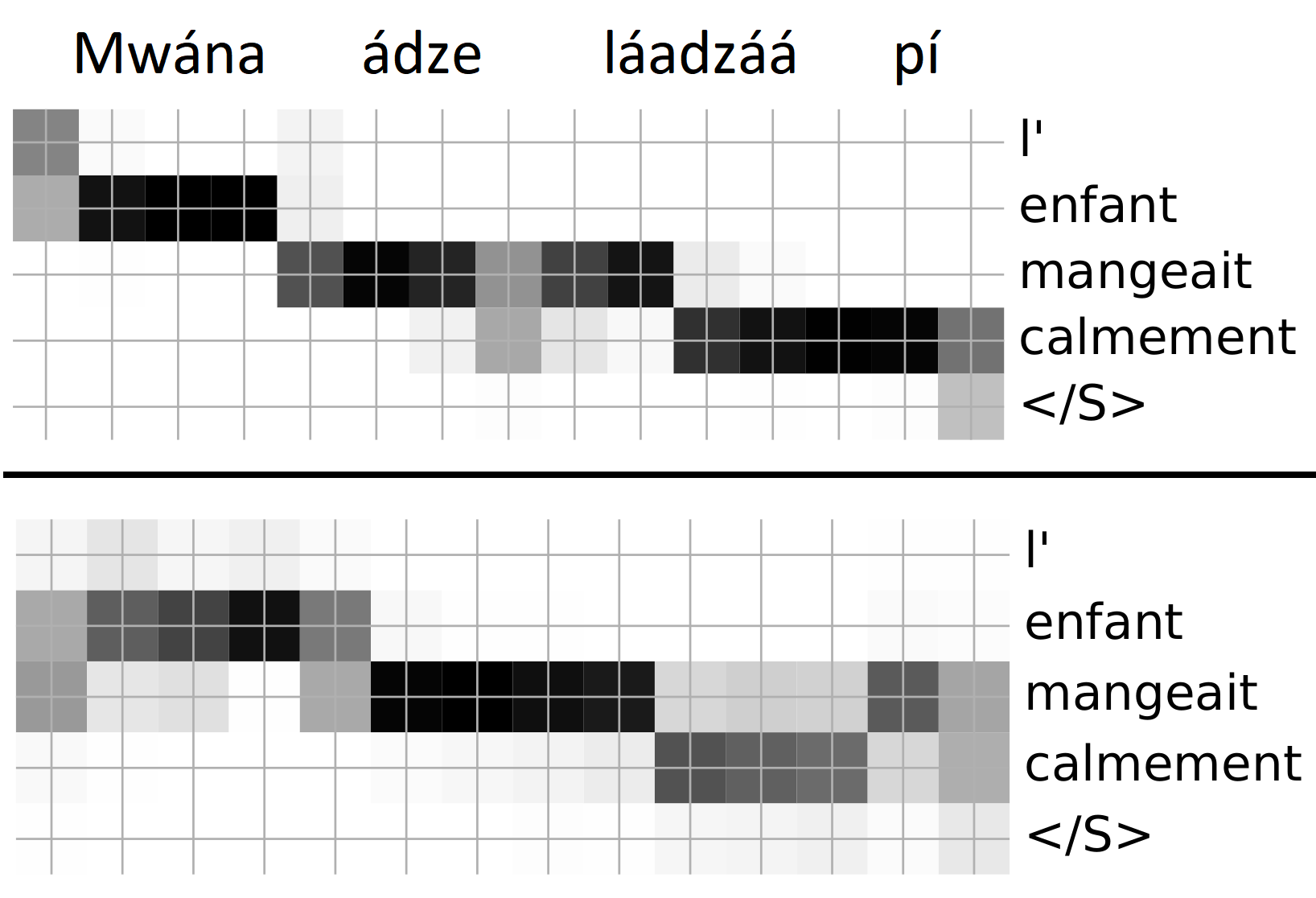}
\caption{NMT output alignment for true phones (top) and AUD using MBN SVAE (bottom). For illustration purposes, we give the transcription of the audio in \embosi{}.}
\label{fig:attmatrix}
\end{figure}

The attention-based segmentation technique remains much more robust for word boundary detection than our monolingual (Bayesian) approach. Figure ~\ref{fig:attmatrix} shows an example of a (good quality)  soft alignment (attention) matrix produced in our best setup (MBN SVAE). 

\section{Related Work}

Word segmentation in a monolingual setup was previously investigated from text input~\cite{Goldwater09bayesian} and from speech ~\cite{aren,lee2015unsupervised,bartels2016toward,elsner2013joint}. Word discovery experiments from text input on Mboshi were reported in~\cite{godard2016preliminary}. 
Bilingual setups (cross-lingual supervision) for word segmentation were discussed 
by~\cite{ip_stueker2008d,stuker:hal-00959225,StahlbergSVS12,boito2017asru}, but 
applied to speech transcripts (true phones).
Looking at NMT from speech, the research by~\cite{berard2016listen, weiss2017sequence} are recent examples of approaches to end-to-end spoken language translation, but using much larger data conditions than ours.

Among the most relevant to our approach are the works of 
\cite{duong2016attentional} on speech-to-translation alignment using attentional NMT and of ~\cite{anastasopoulos2017spoken} for language documentation. However,  the former does not address word segmentation and is not applied to a language documentation scenario, while the latter 
does not provide a full coverage of the speech corpus analyzed. 


\section{Conclusions}

Different from these related works and inspired by \cite{boito2017asru}, this paper presented word segmentation from speech, in a bilingual setup and for a real language documentation scenario (\embosi). 
The proposed approach first performs AUD to generate pseudo-phones from  speech, and then uses these units in an encoder-decoder NMT for word segmentation. 
Our method leads to promising results
for word segmentation from speech, outperforming three baselines in noisy (pseudo-phones) setups and finally delivering the best results reported so far for the Mboshi5k corpus. 
Future work includes investigating sources of weak supervision and minimal viable corpus sizes.
  


\section{Acknowledgements}
This work was partly funded by French ANR and German DFG under grant ANR-14-CE35-0002 (BULB project).
This work was started at JSALT 2017 in CMU, Pittsburgh, and was supported by JHU and CMU (via grants from Google, Microsoft, Amazon, Facebook, Apple). It used the Extreme Science and Engineering Discovery Environment (NSF grant number OCI-1053575 and NSF award number ACI-1445606).

%
%
%

\bibliographystyle{IEEEtran}
\bibliography{mybib}
\end{document}